\documentclass[letterpaper, 10 pt, conference]{ieeeconf}  
\usepackage[colorlinks,urlcolor=blue,linkcolor=blue,citecolor=blue]{hyperref}
\usepackage{color,array}
\usepackage{graphicx}
\usepackage{mathptmx}
\usepackage{algorithm}
\makeatother
\usepackage{mathtools}
\usepackage[utf8]{inputenc}
\usepackage[table]{xcolor}
\usepackage{tabularx,booktabs}
\newcolumntype{C}{>{\centering\arraybackslash}X} 
\usepackage{cite}
\usepackage{amsmath}
\usepackage{amsmath,amssymb,amsfonts}
\usepackage{graphicx}
\usepackage{textcomp}
\usepackage{xcolor}
\usepackage[flushleft]{threeparttable}
\usepackage{hyperref}
\usepackage{caption}
\usepackage{pifont}
\usepackage{xcolor}
\usepackage{rotating}
\usepackage{rotfloat}
\usepackage{algpseudocode}
\usepackage{gensymb}
\usepackage{textcomp}
\usepackage{placeins}
\usepackage{balance}
\usepackage{subfigure}
\usepackage{booktabs}
\usepackage{tabularx}
\usepackage{array}
\usepackage{pbox}
\usepackage{longtable}
\usepackage{booktabs}
\usepackage[mathscr]{euscript}

\DeclareSymbolFont{rsfs}{U}{rsfs}{m}{n}
\DeclareSymbolFontAlphabet{\mathscrsfs}{rsfs}

\def\BibTeX{{\rm B\kern-.05em{\sc i\kern-.025em b}\kern-.08em
    T\kern-.1667em\lower.7ex\hbox{E}\kern-.125emX}}

\IEEEoverridecommandlockouts                              

\title{\LARGE \bf
SPACE: 3D Spatial Co-operation and Exploration Framework for Robust Mapping and Coverage with Multi-Robot Systems
}

\author{Sai Krishna Ghanta \and Ramviyas Parasuraman 
\thanks{School of Computing, University of Georgia, Athens, GA 30602, USA.}
\thanks{Author emails: {\fontfamily{qcr}\selectfont \{sai.krishna;ramviyas\}@uga.edu}}
\thanks{This work is supported by the Army Research Laboratory and was accomplished under  Cooperative Agreement Number W911NF-17-2-0181 (DCIST CRA). 
}
}

\begin{document}

\maketitle
\thispagestyle{empty}
\pagestyle{empty}

\begin{abstract}
In indoor environments, multi-robot visual (RGB-D) mapping and exploration hold immense potential for application in domains such as domestic service and logistics, where deploying multiple robots in the same environment can significantly enhance efficiency. However, there are two primary challenges: (1) the "ghosting trail" effect, which occurs due to overlapping views of robots impacting the accuracy and quality of point cloud reconstruction, and (2) the oversight of visual reconstructions in selecting the most effective frontiers for exploration. Given these challenges are interrelated, we address them together by proposing a new semi-distributed framework (SPACE) for spatial cooperation in indoor environments that enables enhanced coverage and 3D mapping. SPACE leverages geometric techniques, including "mutual awareness" and a "dynamic robot filter," to overcome spatial mapping constraints. Additionally, we introduce a novel spatial frontier detection system and map merger, integrated with an adaptive frontier assigner for optimal coverage balancing the exploration and reconstruction objectives. In extensive ROS-Gazebo simulations, SPACE demonstrated superior performance over state-of-the-art approaches in both exploration and mapping metrics.


\end{abstract}

\section{Introduction}

Multi-Robot Exploration (MRE) is pivotal in advancing robotics research due to its ability to enhance environmental awareness over extended periods, enabling applications such as environmental monitoring, patrolling \cite{espina2011multi}, search and rescue \cite{10611179}, and intelligent transportation \cite{li2018corb}. The core objective of MRE is to synergistically improve the autonomous navigation and mapping capabilities of coordinated robots, optimizing spatial comprehension, cost-effectiveness, travel time, and energy consumption. Recent advancements have seen the development of more efficient and resilient MRE algorithms and systems \cite{almadhoun2019survey}, integrating diverse objectives, sensor modalities, and communication frameworks \cite{latif2024communication}.


Frontier-based MRE methods \cite{latif2024communication}\cite{batinovic2021multi} have received significant attention for their capacity to accelerate exploration by identifying and prioritizing frontiers that offer maximum information gain. However, recent efforts have predominantly focused on building efficient 2D grid maps, optimizing cost, and reducing travel time. These works often fall short when applied to the construction of visual 3D maps. Many existing approaches \cite{lau2022multi,8202319} rely on computer vision-based frontier detection in 2D grid maps, which are not well-suited for the complexities of 3D spatial reconstruction. This highlights a critical gap in the literature, particularly in extracting spatial frontiers and reconstructions.

\begin{figure}[t]
    \centering
    \includegraphics[width = 1\linewidth]{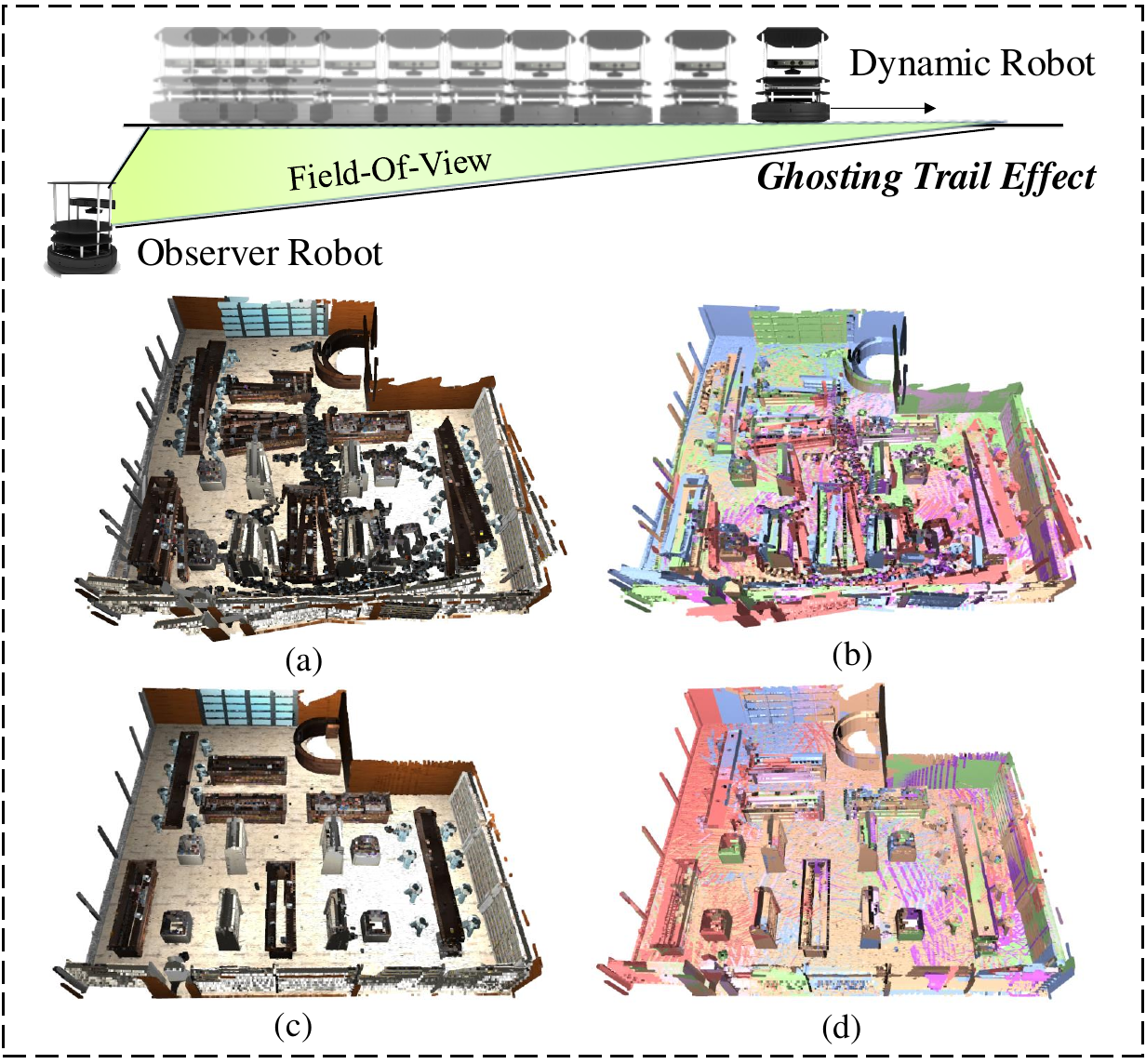}
    \captionsetup{font=small}
    \caption{{\footnotesize{\textbf{The Ghosting Trail Problem:} Formation of ghosting regions \& poor quality spatial maps due to inter-robot visibility during mapping and exploration. (a), (b) and (c), (d) represents explored and composite (where each color represents a local map by each robot) 3D map using RRT with Kimera-Multi \cite{tian2022kimera} and SPACE with RTABMap \cite{labbe2019rtab}, respectively.}}}
    \label{fig:ghost-intro}
    \vspace*{-7mm}
\end{figure}

Compared to collaborative LIDAR-based exploration and large-scale visual (RGB-D camera-based) exploration frameworks, the exploration of visual MRE in indoor environments remains underexplored. This is particularly important because robot interactions and overlapping views in indoor spaces introduce unique dynamic challenges. State-of-the-art (SOTA) SLAM approaches, such as RTAB-Map \cite{labbe2019rtab}, Kimera-Multi \cite{tian2022kimera}, ORB-SLAM3 \cite{campos2021orb}, Swarm-SLAM \cite{lajoie2023swarm}, and CORB-SLAM \cite{li2018corb}, have demonstrated and improved mapping and data communication efficiency in indoor and outdoor environments using improved use of RGB-D data, but inter-robot visibility (appearance) in map reconstructions is typically neglected. These methods often underperform in co-located multi-robot environments, suffering from a "ghosting trail" effect due to overlapping robot exploration paths  (see Fig. \ref{fig:ghost-intro} for an illustration). This ghosting significantly degrades 3D map quality, particularly during critical stages like map merging and frontier identification, resulting in poor spatial mapping and suboptimal exploration. 
These resulting 3D maps can produce unexpected outcomes when used in higher-level semantic classification, clustering, navigation, and decision-making algorithms \cite{ravipati2024object,de2017skimap,ran2021rs,kannan2020material}.

To address this gap, this paper presents a semi-decentralized framework for multi-robot 3D spatial exploration and mapping, termed SPACE. Our approach enhances the accuracy of generating dense 3D metric mesh models by introducing mutual awareness, enabling robots to operate collaboratively in shared environments. In addition, we propose a bi-variate spatial frontier detection method, a dynamic robot filter, and a coherent spatial map merger for superior multi-robot mapping. We further introduce an adaptive frontier assigner that optimizes spatial information gain and the quality of dense metric map construction and achieves an optimal exploration, improving the 3D reconstruction accuracy and spatial coverage performance.

The core novelties and contributions of this paper are 
\begin{itemize}
    \item The introduction of geometric-based mutual awareness and dynamic robot filter methods to address the spatial constraints in visual mapping, significantly improving 3D multi-robot mapping in indoor environments. 
    \item A novel MRE approach, leveraging frontier importance balancing the exploration and 3D reconstruction objectives and an adaptive exploration validator to optimize exploration efficiency and coverage.
\end{itemize}
We demonstrate the effectiveness of SPACE in comprehensive simulation experiments and validate against state-of-the-art multi-robot mapping methods such as RTAB-MAP \cite{labbe2019rtab}, Kimera-Multi \cite{tian2022kimera} and exploration approaches such as RRT \cite{zhang2020rapidly}, DRL \cite{hu2020voronoi}, and SEAL \cite{latif2023seal}. 
Finally, we open-source SPACE\footnote{\url{https://github.com/herolab-uga/SPACE-MAP}} as a ROS package to facilitate its adoption and further development by the broader robotics community. A video supplement demonstrating the approach in simulations and real-world robots is available\footnote{\url{https://youtu.be/EE0velFrJgI}}.

\section{Related Work}
Frontier-based exploration is widely used in MRE, where the robot works towards maximizing its exploration by moving toward the unexplored areas on the map. Existing works \cite{yamauchi1997frontier} \cite{yamauchi1998frontier} aim greedily to push robots either to the closest frontier or most uncertain region for maximization of coverage. Rapidly-exploring Random Trees (RRTs) \cite{zhang2020rapidly} have significantly performed in planning multi-robot exploration schemes for fast and efficient exploration. However, they suffer from suboptimal solutions due to the stochastic nature of RRTs. Deep Reinforcement Learning (DRL) approaches have been proposed to enhance multi-robot exploration. In \cite{hu2020voronoi}, DRL is integrated with Voronoi-based cooperative exploration and was limited by training challenges to distribute the coverage for each robot. SEAL \cite{latif2023seal}, a recent Gaussian Processes-based information fusion framework, is introduced to maximize the efficiency of exploration and localization. However, this approach fails to tackle non-linearized hulls with their convex hull optimization. 
In CQLite \cite{latif2024communication}, an enhanced Q-learning was introduced to solve the problems of revisiting areas that are already explored by the robots. 
Nonetheless, the existing approaches rely on contour-based and feature-based techniques for frontier detection \cite{keidar2014efficient} and map merging \cite{horner2016map}. These frontier detection algorithms are often sensitive to noise. The feature-based map merging provides a fuzzy grid map and highly depends on the robots' initial positions. 

Most MRE approaches rely on the grid map for frontier detection, goal assignment, and navigation. The Visual Simultaneous Localization and Mapping (VSLAM) approaches \cite{labbe2019rtab,mur2015orb} aid the MRE with spatial mapping and are effective for single-robot exploration in static environments. 
The semantic VSLAMs \cite{krishna20233ds} are introduced to address the issues, primarily the ghosting effect, in dynamic environments caused by humans. Semantic VSLAMs eliminate dynamic features by detecting humans using semantic segmentation or object detection. However, they are ineffective for multi-robot environments because of the rapid dynamic movements in robots and the difficulty in training the recognition algorithms for all kinds of robots, unlike humans.  

Contrary to the existing works, SPACE addresses and overcomes the challenges associated with MRE in visual 3D maps. SPACE maximizes the efficiency of the 3D map with a novel robot dynamic filter and map merging algorithm. The exploration strategy is designed to consider the 3D information gain, which balances the unexplored and weakly reconstructed frontiers for effective spatial exploration. 

\section{Proposed Methodology}
\label{sec:methodology}
\begin{figure*}[t]
    \centering
    \captionsetup{font=small}
    \includegraphics[width = 0.95\linewidth]{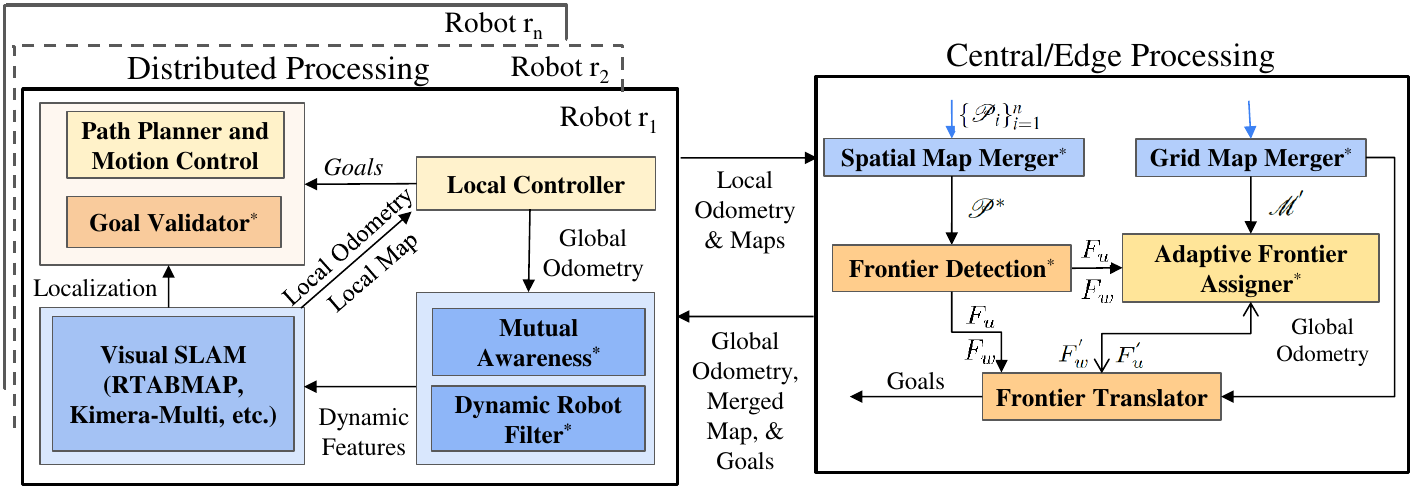}
    \caption{{\footnotesize{Overview of the proposed methodology. The \textcolor{blue}{Blue}-shaded components are \textbf{Mapping-related} modules, and the \textcolor{orange}{Orange}-shaded components are \textbf{Exploration-realated} modules. Marked with an asterisk(*) are novel elements introduced in this paper for multi-robot exploration.}}}
    \label{fig:overview}
    \vspace{-4mm}
\end{figure*} 

\noindent \textbf{Problem Setting:}
Let there be $n$ robots denoted by the set $R = \{r_1, r_2, \dots, r_n\}$, each with known initial position in a global frame. Each robot $r_i$ maps the environments, creating a spatial map $\mathcal{P}_{i}$. All these spatial maps $\{\mathcal{P}_1, \mathcal{P}_2, \dots, \mathcal{P}_n\}$ are merged to create a global exploration map $\mathcal{P}^*$. The frontiers $F$ identified in $\mathcal{P}^*$ are categorized into two sets: unexplored $F_u$ and weakly explored $F_w$ based on their densities and variances, respectively. The frontier assigner assigns a frontier $f_a$ with maximum revenue value in $U(r_i, \{F_u, F_w\} \in F)$ to each robot $r_i$. To minimize the latency within the resources, the streaming updates of the local and merged maps are computed and communicated by edge/central processing.

The SPACE is a semi-distributed framework with centralized map merging, frontier detection, and frontier assignment modules, as shown in Fig. \ref{fig:overview}. It consists of on-board (distributed) processes such as Visual SLAM, mutual awareness, dynamic robot filtering, frontier validation, and path planning. Moreover, SPACE utilizes translated frontiers $F_u^{'}$, $F_w^{'}$ within 2D grid map $\mathcal{M}^{'}$ for path planning and navigation. SPACE can be integrated with existing visual SLAMs (RTABMap \cite{labbe2019rtab} and Kimera-Multi \cite{tian2022kimera}) for localization and RGB-D mapping. Though 2D grid map $\mathcal{M}^{'}$ can be extracted from the visual SLAMs, existing 2D SLAM approaches such as gmapping \cite{grisetti2007improved} and SLAM toolbox \cite{macenski2021slam} can be integrated.

\subsection{Mutual Awareness Module}
The proposed mutual awareness is a geometric-based approach to determine whether the other robots are within the Field-of-View (FoV) of the observer robot. Moreover, this approach provides an ability to recognize the presence of other robots within their FoV \& avoid explicit visual mapping. This method performs pairwise iteration with all the robots in the environment. Consider an observer and a target in a global frame, with their respective pose $\mathbf{p}_{\text{o}} = (x_{\text{o}}, y_{\text{o}},\psi_{\text{o}} ),  \mathbf{p}_{\text{t}} = (x_{\text{t}}, y_{\text{t}}, \psi_{\text{t}})$. We solve the problem on the plane, ignoring the $z$ component, roll, and pitch in the pose. Firstly, we detect whether any other robot is within the RGB-D mapping range before FoV calculation. This effective detection of robot proximity by the sensor is valid and further processed only if \( 0 \leq \|\mathbf{p}_o - \mathbf{p}_t\| \leq \gamma \), where \( \gamma \) is the maximum range of the RGB-D camera. If any target robot is within proximity, we calculate the angle $\theta$ between the observer and target. 
To ensure the angles are in between $[- \pi, \pi]$, we perform angle normalization of observer yaw $\psi$ and $\theta$ as 
\begin{equation}
\begin{aligned}
    \theta_{\text{norm}} = \left((\theta + \pi) \mod 2\pi\right) - \pi, \\
    \psi_{\text{norm}} = \left((\psi_{\text{obs}} + \pi) \mod 2\pi\right) - \pi.
\end{aligned}
\end{equation}

The normalized angular difference between the observer's orientation and the vector pointing to the target is computed by
$\Delta \theta = \left(\theta_{\text{norm}} - \psi_{\text{norm}} + \pi\right) \mod 2\pi - \pi$.
Considering robots as circular objects maximizes precision in awareness. The target robot with its largest dimension as the radius is represented as angular size \(\alpha = \arctan\left(\frac{R}{\|\mathbf{p}_o - \mathbf{p}_t\|}\right) \), as viewed from the observer. The visibility from the observer's camera can then be estimated by comparing the angle \(\Delta \theta\) to the effective half FoV and angular size of the robot. 
The target robot is within the observer robot's FoV if:
$|\Delta \theta| + \frac{c}{\|\mathbf{p}_o - \mathbf{p}_t\|} \leq \frac{\text{FoV}_{\text{cam}}}{2} -  \alpha 
$, where \(\text{FoV}_{\text{cam}}\) is the horizontal FoV of the camera, 
$R$ is the robot's radius, and $\frac{c}{\|\mathbf{p}_o - \mathbf{p}_t\|}$ is the dynamic buffer to balance the latency with distance.  

\subsection{Dynamic Robot Filter (DRF) Module}
When the target robot is within the FoV, the observer robot continuously maps the target robot, leading to a ghosting trail effect in the 
map. The proposed DRF eliminates the dynamic features of the target robot while mapping. The DRF estimates the target robot's position in the image coordinate frame. 
Converting a world coordinate system to an image coordinate system is often challenging. It involves precise utilization of both intrinsic \& extrinsic camera parameters, and real-world camera pose to translate the coordinates. 

Given a robot observer and a target, the objective is to determine the image coordinates \((u, v, d)\) on the observer's camera at which the target appears. This involves translating the target's position from the world coordinate system to the camera's image plane through a series of transformations. Let the observer's camera height from the ground be denoted by \(h_{\text{o}}\). First, we calculate the relative position of the target with respect to the observer in a 3D coordinate system from a 2D coordinate system with \(h_{\text{o}}\). Later, we calculate the relative position vector to get the position in the observer's frame as 
\begin{equation}
\mathbf{p}_{\text{obs-frame}} = \mathbb{R}(-\psi_{\text{obs}}). (\mathbf{p}_{\text{rel-2D}}[0], h_{\text{obs}}, \|\mathbf{p}_{\text{rel-2D}}\|)^T ,
\end{equation}
where $\mathbb{R}(.)$ is the rotation matrix constructed to account for the observer's yaw $\psi$ in the plane, and $\mathbf{p}_{\text{rel-2D}} = \mathbf{p}_{\text{t}} - \mathbf{p}_{\text{o}}$ refers to the position of relative position in 3D co-ordinate system $\mathbf{p}_{\text{rel-3D}}$. The $\mathbf{p}_{\text{obs-frame}}$ in the coordinates of the target relative to the observer, rotated and adjusted for height, giving the observer's perspective in 3D. After the estimation of position of target in observer's frame, the position is projected to the image plane with the intrinsic and extrinsic camera parameters. The intrinsic camera matrix, which transforms 3D camera coordinates into 2D image coordinates, is represented as \( \mathbf{K} \). The extrinsic parameters $\mathbf{M}_{\text{ext}}$ are encapsulated by the rotation matrix \( \mathbf{R}_{\text{ext}} \) and the translation vector \( \mathbf{T}_{\text{ext}} \), which align the camera with respect to a global reference frame.
The matrix product of camera parameters with the homogeneous coordinates of the target's position in the observer's frame yields normalized image coordinates as shown:
\begin{equation}
    \begin{bmatrix}
u \\
v \\
d
\end{bmatrix} = \mathbf{K} \left[ \mathbf{M}_{\text{ext}}; \begin{array}{c}
0 \\
0 \\
0 \\
1
\end{array} \right] \cdot \begin{bmatrix}
\mathbf{p}_{\text{obs-frame}} \\
1
\end{bmatrix}
\end{equation}

Moreover, a 3D bounding box with dimension $\frac{c}{\|\mathbf{p}_o - \mathbf{p}_t\|}$, where $c$ is an upscaling constant, is introduced centering $(u, v, d)$ to extract the region of target robot within the observer frame. A DBSCAN clustering algorithm segments the robot within the bounding box and eliminates the dynamic features within the cluster.

\subsection{Coherent Spatial Map Merger Module}
The objective of spatial map merger is to find a set of rigid transformations \(\{T_i\}_{i=1}^n\) that align all point clouds \(\{\mathcal{P}_i\}_{i=1}^n\) into a unified, coherent global coordinate space. This approach aims to minimize the global alignment error across transformations of point clouds simultaneously to calculate the merged map $\mathcal{P}^*$ with maximum alignment:
\begin{equation}
\min_{\{T_i\}} \sum_{(i,j) \in \mathcal{E}} \Lambda_{i,j}  \cdot \| T_i \mathcal{P}_i - T_j \mathcal{P}_j \|^2 .
\end{equation}
Here, \( T_i \) and \( T_j \) are the transformations applied to point clouds \( \mathcal{P}_i \) and \( \mathcal{P}_j \) respectively, \( \mathcal{E} \) represents set of all edges in the interconnected pose graph consisting point clouds \(\{\mathcal{P}_i\}_{i=1}^n\) as nodes. Each edge, for instance $\mathcal{P}_i$ to $\mathcal{P}_j$, in pose graph is quantified using transformation matrices \( T_{i,j} \) and information matrices \( \Lambda_{i,j} \). The transformation matrix \( T_{i,j} \) aligns \( \mathcal{P}_i \) to \( \mathcal{P}_j \), and the information matrix \( \Lambda_{i,j} \) quantifies the certainty of this alignment.  This pose graph is responsible for all the simultaneous transformations of the point clouds. The existing SOTA approaches, Kimera-Multi \cite{tian2022kimera}, do not consider the non-linearities in transformation estimation, which results in noise and sub-optimal solutions. The Levenberg-Marquardt algorithm \cite{more2006levenberg} is utilized to solve this non-linear least squares problem. This method interpolates between the Gauss-Newton algorithm and gradient descent, offering a robust approach for dealing with non-linearities in the iterative transformation estimation:
\begin{equation}
T^{(k+1)} = T^{(k)} - (J^T \Lambda J + \mu I)^{-1} J^T \Lambda r,
\end{equation}
where \( J \) is the Jacobian matrix of the residuals \( r \) with respect to the transformations.
\( \Lambda \) is a block-diagonal matrix consisting of all information matrices \( \Lambda_{i,j} \), and \( \mu \) is the damping factor that adjusts the algorithm's convergence behavior. The iterations continue until the alignment error function reaches its global minimum.

\subsection{Bi-variate Spatial Frontier Detection  Module}
\begin{figure*}[t]
    \centering
    \includegraphics[trim={0 6mm 0 0},clip,width = \linewidth]{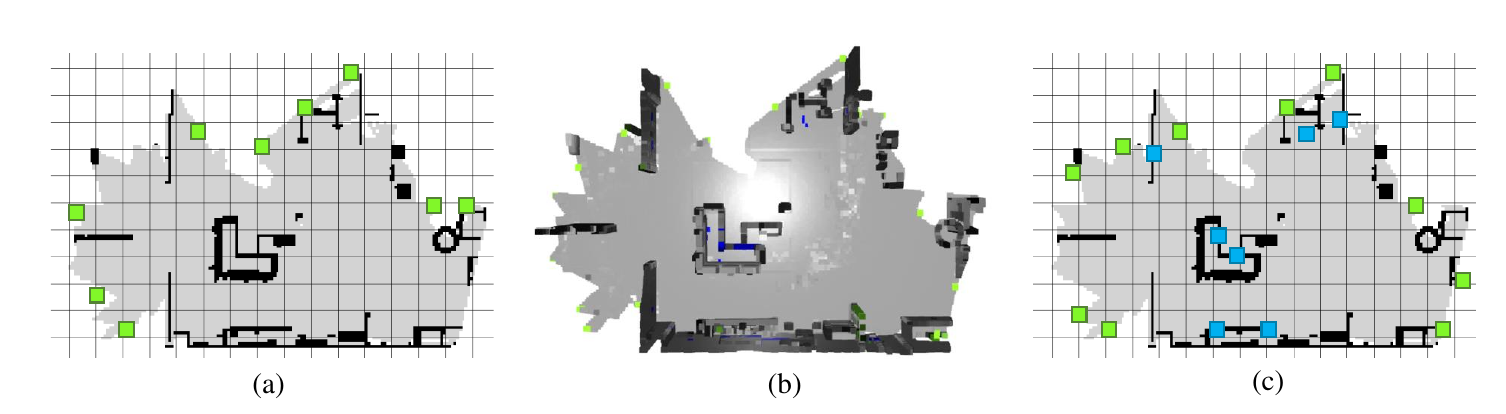}
    \caption{(Left) Contour-based Frontier Detection \cite{keidar2014efficient} (Center) Bi-Variate Spatial Frontier Detection of SPACE (Right) Grid map with Translated Spatial Frontiers.}
    \label{fig:frontier}
    \vspace{-4mm}
\end{figure*}

In recent works, frontiers in an environment are typically delineated by contours along the edges in grid maps. However, these frontiers do not account for partially constructed \& unexplored regions in a 3D space. We propose a novel approach for identifying two-classes of frontiers: 1.) weakly explored regions 2.) unexplored regions within dense-metric 3D maps as shown in Fig. \ref{fig:frontier}. These frontiers are detected within the merged map $\mathcal{P}^*$, where each point \( p_i \in \mathcal{P}^* \) is a vector in \( \mathbb{R}^3 \). We define a voxel grid with side length \( \varrho \) to obtain downsampled point cloud, \( \mathcal{P}^{*}_{d} \). The density $\rho(p_i)$ at a point \( p_i \) is calculated based on the number of neighbor points \(p_j\) within a specified radius \( r_{s} \):
\begin{equation}
\rho(p_i) = |\{ p_j \in \mathcal{P}^{*}_{d}  : \|p_j - p_i\| \leq r_{s} \}|
\label{eqn:density}
\end{equation}
The variance \( \sigma^{2}(p_i) \) for each point \( p_i \) in a point cloud is computed to calculate the spatial variability among the set of neighboring points \( N_{\rho}(p_i) \) within the radius \( \rho \).
The density $\rho(p_i)$ and variance $\sigma^{2}(p_i)$ are calculated efficiently using a KD-Tree to avoid a brute-force search through all points. The density of a point presents the unexplored region, whereas the variance represents the weakly constructed regions. The unexplored points  \textbf{$U$} are identified by densities below a threshold \(\delta_{\textit{d}}\), and weakly constructed indices \textbf{$W$} by variances exceeding the threshold \(\delta_{\textit{v}}\), as shown in Alg. \ref{algof}.

\vspace{-2mm}
\begin{algorithm}[h]
\caption{3D Frontier Detection in Point Cloud}
\label{algof}
\begin{algorithmic}[1]
\Require $ \mathcal{P}^{*}$, $\varrho$, $r_s$, $\delta_{\text{d}}$, $\delta_{\text{v}}$ 
\Ensure $F_u, F_w$
\State $\mathcal{P}^{*} \leftarrow$ Downsample($\mathcal{P}$, $\varrho$)
\State Initialize arrays: \textit{$\rho$, $\sigma^2$}
\For{each point $p_i \in \mathcal{P}^{*}_{d}  $}
    \State $\rho(p_i)$ $\leftarrow |\{p_j \in \mathcal{P}^{*}_{d}  : \|p_i - p_j\| \leq \rho\}|$
    \State  $N_{\rho}(p_i)$  $\leftarrow \{p_j \in \mathcal{P}^{*}_{d}   : \|p_i - p_j\| \leq \rho\}$
    \State $\overline{p} \leftarrow \frac{1}{|\text{$N_{\rho}(p_i)$}|} \sum_{p_j \in \text{$N_{\rho}(p_i)$}} p_j$
    \State $\sigma^{2}(p_i)$ $\leftarrow \frac{1}{|\text{$N_{\rho}(p_i)$}|} \sum_{p_j \in \text{$N_{\rho}(p_i)$}} \|p_j - \overline{p}\|^2$
\EndFor
\State \Return $F_u$ $\leftarrow \{i : \rho(p_i)< \delta_{\text{d}}\}$, 
$F_w$ $\leftarrow \{i : \sigma^2(p_i) > \delta_{\text{v}}\}$
\end{algorithmic}
\end{algorithm}

\subsection{Adaptive Frontier Assignment Module}
This module is responsible for the mathematical and operational dynamics of the frontier assignment within the SPACE exploration framework. We categorized the entire mechanism into two parallel threads:

\noindent \textbf{3D Information Gain:} It describes the possibility of gaining new information through frontier exploration. The  set of voxels $V_f$ within the region around the frontier $f$ are considered in estimation of possible information gain. The information gain \( I_g(f) \) for a frontier $f$ is the average non-probabilistic Shannon entropy \cite{wuest2003information} across its constituent points, as shown in the equation below:
\begin{eqnarray}
    I_g(f) = 
\begin{dcases} 
 - \frac{1}{|V_f|} \sum_{v \in V_f} \rho(v) \log(\rho(v))
 & \text{if } f \in F_u \\
 - \frac{1}{|V_f|} \sum_{v \in V_f} \sigma^{2}(v) \log(\sigma^{2}(v)) 
 & \text{if } f \in F_w 
\end{dcases}
\label{eqn:frontier-gain}
\end{eqnarray}

\noindent \textbf{Frontier Importance:} 
The frontier identifier categorizes the frontiers into $F_u$ and $F_w$. Initially, exploration prioritizes the unexplored regions, but over time, the importance shifts towards weakly constructed regions. We proposed an important function that considers the class of frontier, time of exploration, and distance to frontiers. The distance factor $\mathscr{D}$ represents the average Euclidean distance from the centroid $c$ at $l_c$ of all robots to the nearest unexplored frontier, relative to the average distance to weakly explored frontiers ($\mathscr{D} =  \frac{\sum_{f \in F_u} \|l_c - f\| }{\sum_{f \in F_w} \|l_c - f\|}$). The time factor $T(t)$ measures elapsed time relative to the start of the exploration phase. The frontier importance $I_t(f)$ is expressed as:
\begin{eqnarray}
    I_t(f) = 
\begin{dcases} 
 \frac{e^{-\lambda T(t)}  \times \mathscr{D}}{1 + e^{-\lambda (T(t) - \xi)}} 
 & \text{if } f \in F_u , \\
1 - \frac{e^{-\lambda T(t)} \times \mathscr{D}}{1 + e^{-\lambda (T(t) - \xi)}}
 & \text{if } f \in F_w ,
\end{dcases}
\label{eqn:frontier-importance}
\end{eqnarray}
where $\lambda $ is a scaling parameter and \( \xi \) is a threshold parameter that determines the transition in the sigmoid function. The revenue function of a frontier with respect to a robot $r_i$ at position $\mathbf{p}_{i}$ is calculated with 3D information gain, frontier importance and heuristic distance $h$ between $\mathbf{p}_i$ and translated frontier $\mathbf{p}_{f^{'}}$ with $\kappa$ is a scaling constant, as shown below.
\begin{equation}
    U(r_i, f^{'}) = \frac{{I}_t(f) \cdot I_g(f)}{\kappa \times h(\mathbf{p}_i, \mathbf{p}_{f^{'}})} ,
\end{equation}
The maximum revenue value within $U(r_i,F^{'}) \in \{U(r_i,f^{'}_1), U(r_i,f^{'}_2)...U(r_i,f^{'}_m)\}$ is assigned to the robot $r_i$. 

\subsection{Adaptive Exploration Goal Validator Module}
Generally, certain robots, especially UGVs, are not capable of mapping the entire environment due to their 2D mobility and height, resulting in an increase in invalid frontiers (which cannot be explored). To tackle uncertain or invalid frontiers, which significantly extend the duration of robot explorations, we introduce the adaptive exploration validator, which estimates the exploration duration effectively. To facilitate this, we employ the A* path planning algorithm to determine the shortest path \( P \) between the 2D translated frontier \( f^{'} \) and the robot's current position \( \mathbf{p} \). The exploration time \( \tau \) required to traverse the sequence of points on the path $\mathbf{P} = \{\mathbf{p_1}, \mathbf{p_2}, \ldots, \mathbf{p_{\tau}\}}$ (for a fixed time horizon $\tau$) is calculated as follows: 
\begin{equation}
t_e(\mathbf{p}, \mathbf{p_{f^{'}}}) = t_e(\mathbf{p}, {\mathbf{p_{1}}}) + t_e(\mathbf{p_{1}},\mathbf{p_{2}}) + \ldots + t_e(\mathbf{{p}_{n}}, \mathbf{p_{f^{'}}})
\end{equation}

The Adaptive Exploration Validator monitors the exploration time for the last $\beta$ points within the  $n$-point estimated trajectory. If the robot exploration time is greater than the $t_e(R_\beta, R_f)$, then the frontier $f$ is considered as invalid.


\subsection{Algorithmic Complexity}
The overall time complexity of the Mutual Awareness is $O(n_{p} \times n)$, and is dependent on the number of robots ($n_{p} < n)$ within proximity. The time complexity of DRF, Spatial Frontier Detection, 3D information gain, and Frontier Importance are $O(1)$, $O(n^2)$, $O(n)$, and $O(n \times m)$, respectively. Here, $n$ is the number of robots, and $m$ is the number of frontiers.

\begin{figure*}[t]
    \centering
    \captionsetup{font=small}
    \includegraphics[width = \linewidth]{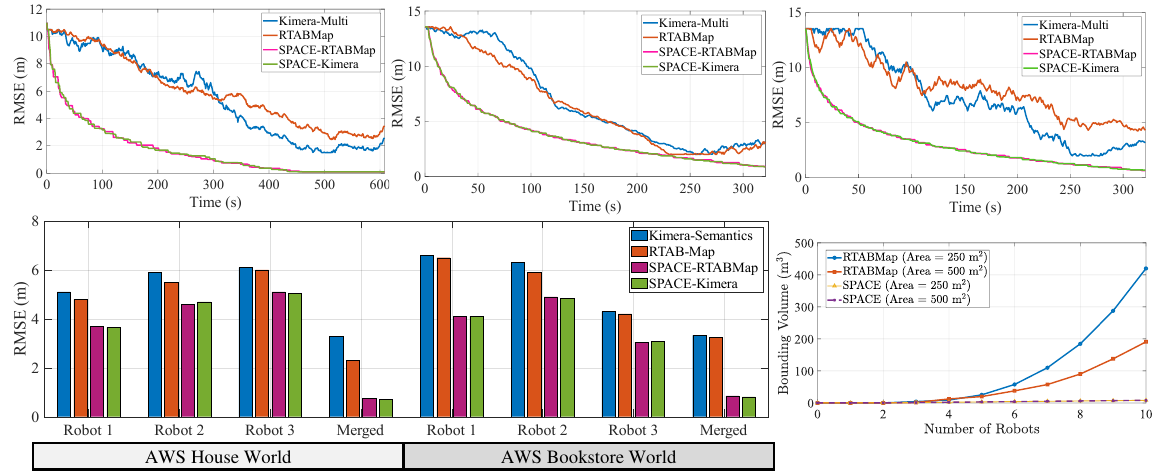}
    \caption{\footnotesize{Performance of the spatial mapping in multi-robot exploration. Here, we used the SPACE exploration approach described in Sec.~\ref{sec:methodology} in all mapping variants for a fair comparison. The top row subplots show the 3D reconstruction accuracy in the three scenarios tested (\textbf{(Left)}: AWS House World (3 Robots), \textbf{(Middle)}: AWS Bookstore World (3 Robots), and \textbf{(Right)}: AWS Bookstore World (6 Robots)). In the bottom row, the \textbf{(Left)} plot shows the detailed robot-wise performance variations, and \textbf{(Right)} plot shows the impact of the ghosting trail effect (bounding volume of the reconstruction inaccuracies) by increasing the robot density in a given area. SPACE-RTABMap is invariant to the ghosting effect with an increase in density.}}
    \label{fig:3dplotsresults}
    \vspace{-2mm}
\end{figure*}


\section{Experimental Results and Analysis}
We implemented the approach in the ROS-\textit{Gazebo} framework and Turtlebot-3 Waffle Robots. 
SPACE is employed with the ROS \textit{RTAB-Map} package \cite{rtabmap_ros} \& \textit{Kimera-Multi} \cite{tian2022kimera} for visual SLAM within each robot, generating 2D \& 3D submaps using data from the camera sensor. The 2D sub-maps, utilized for path planning, are merged by \textit{multirobot-map-merge} \cite{multirobot_map_merge} without considering the initial positions of robots. The ROS \textit{move-base} package \cite{move_base} enables the robots to navigate toward their designated goals while avoiding obstacles.
This package incorporates the A* algorithm for long-range path planning and the Dynamic Window Approach (DWA) \cite{fox1997dynamic} for real-time obstacle evasion. The global frame refers to a reference frame fixed at a point in the world, while the local frame pertains to a specific robot's reference frame. 


We conducted experiments in two indoor simulation environments: AWS House \cite{aws_house} (70$m^{2}$ area) and AWS Bookstore \cite{aws_bookstore} (100$m^{2}$ area). Robots are mounted with Kinect-V2 RGB-D camera with horizontal $\text{FoV}_{cam}=84.1$\degree and maximum range $\gamma=5m$ for gathering the spatial information. 
Evaluations focused primarily on three distinct setups to test the system's effectiveness and adaptability: three robots navigating the house, three robots in the bookstore, and six robots operating within the bookstore. 
Robots had a max. linear speed of \(0.5 \, \text{\textit{m/s}}\) and a turning rate of \(\frac{\pi}{4} \, \text{\textit{rad/s}}\).

The ground truth maps, for experimental analysis, are generated with the \textit{Gazebo Map Generator} \cite{gazebo_map_creator}. For evaluation, the explored and ground-truth meshes are sampled uniformly with \(10^3 \, \text{points/m}^2\) as in \cite{tian2022kimera}.

\noindent \textbf{Performance Metrics:}
The metrics below are used to evaluate mapping and exploration objectives comprehensively.
\begin{itemize}
    \item \textbf{Bounding Volume (BV)}: The volume of the 3D map 
    created by the ghosting trail effect.
    \setlength\belowcaptionskip{-20pt}
    \item \textbf{Total Coverage (2D/3D)}: The \% of a map in the 2D (grid) or 3D (point cloud) that has been mapped relative to the respective ground truth. The 3D total coverage is calculated after the removal of the ghosting region. 
    \item \textbf{Overlap Percentage (2D/3D)}: The percentage of a map in the 2D (grid) or 3D (point cloud) that has been covered more than once during the mapping process.
    \item \textbf{Mapping Time}: To assess efficiency, the total duration (s) for exploration averaged across the trials is reported.
    \item \textbf{Total Distance}: The cumulative distance (m) (i.e., path length) traveled by all the robots during the trial.
\end{itemize}

\subsection{Mapping Performance}

\begin{table}[t]
\centering
\scriptsize
\captionsetup{font=small}
\caption{\footnotesize{Performance comparison of SPACE with other base mapping methods in the AWS Bookstore world with 3 Robots (Note: SPACE's exploration approach is used for all the compared methods). Best values are in \textbf{bold} face and asterisk($^{*}$). The results highlight the benefits of our DRF layer in improving 3D reconstruction accuracy and its effect in exploration.}}
\label{tab:performance-mapping}
\begin{tabular}{|p{2.5cm}|p{1.3cm}|p{1.3cm}|p{1.3cm}|p{1.7cm}|}
\hline
\textbf{Metrics} & \textbf{RTAB-Map} & \textbf{Kimera-Multi} & \textbf{SPACE  w/ RTABMap} & \textbf{SPACE   w/ Kimera-Multi} \\ \hline
Mapping Time (s) & $359 \pm 51$ & $338 \pm 64$ & $\mathbf{321 \pm 21}^*$ & $324 \pm 23$ \\ \hline
Total Distance (m) & $247 \pm 18$ & $253 \pm 12$ & $\mathbf{142 \pm 8}^*$ & $152 \pm 6$ \\ \hline
2D Coverage (\%) & $93 \pm 2$ & $95 \pm 4$ & $100 \pm 3$ & $\mathbf{100 \pm 2}^*$ \\ \hline
3D Coverage (\%) & $87 \pm 4$ & $90 \pm 6$ & $91 \pm 3$ & $\mathbf{94 \pm 6}^*$ \\ \hline
2D Overlap (\%) & $42 \pm 8$ & $43 \pm 3$ & $\mathbf{35 \pm 1}^*$ & $37 \pm 4$ \\ \hline
3D Overlap (\%) & $41 \pm 1$ & $47 \pm 8$ & $\mathbf{38 \pm 2^*}$ & $40 \pm 2$ \\ \hline
3D-Map RMSE (\%) & $2.01$ & $2.76$ & $\mathbf{0.12}^*$ & $0.14$ \\ \hline
BV (\%) & $9.1$ & $8.86$ & $\mathbf{0.01}^*$ & $0.015$ \\ \hline
\end{tabular}
\end{table}

The performance of the spatial mapping during and after MRE in different scenarios are depicted in Table \ref{tab:performance-mapping} and Fig. \ref{fig:3dplotsresults}. 
We tested the SPACE framework integrated with recent VSLAMs such as RTAB-Map \cite{labbe2019rtab} and Kimera-Multi \cite{tian2022kimera}, in addition to standalone evaluations of RTAB-Map and Kimera-Multi, to facilitate a clear comparison. The SPACE-integrated VSLAMs are equipped with our mutual awareness, dynamic robot filter, and map merging modules, whereas standalone approaches are equipped only with the SPACE exploration approach and ICP map merging as in Kimera-Multi \cite{besl1992method}.  Unlike the RMSE trend observed with RTAB-Map and Kimera-Multi, which fluctuated, the RMSE of the SPACE, when integrated with these systems, exhibited a more consistent and stable decrease throughout exploration. This is because of the generation of ghosting regions throughout the exploration in the standalone systems. 

The overall performance of the SPACE with both RTAB-Map and Kimera-Multi is 25 times higher than the standalone RTAB-Map and Kimera-Multi. As the number of robots increases, the RMSE tends to be more unstable throughout the exploration due to the increase in the frequency of the ghosting effect. Moreover, while SPACE shows a slightly lower RMSE (-24.8\%) compared to individual robot-wise map comparisons, it outperforms in the RMSE of merged maps, demonstrating superior performance when integrating spatial maps from multiple robots. Moreover, the experiments with isolated environments depicted an exponential relationship between the number of robots and the bounding volume of the ghosting region. This implies the advantages of the SPACE are amplified in a denser robot deployment.

\begin{table*}[ht]
  \centering
  \scriptsize
\captionsetup{font=small}
  \caption{\footnotesize{Performance comparison of SPACE with other exploration methods. Note: RTAB-Map is used as the 3D mapping layer for all methods. 
  }}
  \label{tab:explo-com}
  \resizebox{\textwidth}{!}{%
  \begin{tabular}{|l|c|c|c|c|c|c|c|c|c|c|c|c|}
    \hline
    \textbf{Evaluation parameters} & \multicolumn{4}{c|}{\textbf{Three robots in AWS House world}} & \multicolumn{4}{c|}{\textbf{Three robots in AWS Bookstore world}}& \multicolumn{4}{c|}{\textbf{Six robots in AWS Bookstore world}} \\
    \hline
    & \textbf{RRT \cite{zhang2020rapidly}} & \textbf{DRL \cite{hu2020voronoi}} & \textbf{SEAL \cite{latif2023seal}}  & \textbf{SPACE} & \textbf{RRT \cite{zhang2020rapidly}} & \textbf{DRL \cite{hu2020voronoi}} & \textbf{SEAL \cite{latif2023seal}}  & \textbf{SPACE} 
    & \textbf{RRT \cite{zhang2020rapidly}} & \textbf{DRL \cite{hu2020voronoi}} & \textbf{SEAL \cite{latif2023seal}} & \textbf{SPACE} \\
    \hline
    Mapping Time (s) & $608 \pm 52$ & $466 \pm 67$ & $\mathbf{457 \pm 34^{*}}$  & $463 \pm 27$ & $347 \pm 32$ & $324 \pm 21$ & $324 \pm 18$  & $\mathbf{321 \pm 17^{*}}$ & $212 \pm 18$& $267 \pm 29$& $\mathbf{197 \pm 11}^{*}$ & ${199 \pm 9}$ \\ \hline
    Total Distance (m) & $192 \pm 11$ & $104 \pm 19$ & $156 \pm 11$& $\mathbf{154 \pm 9^{*}}$ & $278 \pm 26$ & $235 \pm 9$ & $\mathbf{142 \pm 8^{*}}$ & $151 \pm 18$ & $223 \pm 12$ & $196 \pm 17$ & $141 \pm 18$& $\mathbf{138 \pm 6^{*}}$ \\ \hline
    Total 2D Coverage (\%) & $87 \pm 4$ & $91 \pm 3$ & $90 \pm 2$ &$\mathbf{97 \pm 3}^{*}$ & $90 \pm 5$ & $93 \pm 2$ & $99 \pm 1$ & $\mathbf{100 \pm 3^{*}}$ & $93 \pm 4$ & $\mathbf{94 \pm 5^{*}}$ & $93 \pm 1$ & $93 \pm 2$\\ \hline
    Total 3D Coverage (\%) & $79 \pm 3$ & $88 \pm 2$ & $92 \pm 2$ & $\mathbf{95 \pm 1^{*}}$ & $82 \pm 2$ & $88 \pm 4$ & $91 \pm 3$ & $\mathbf{98 \pm 1^{*}}$ & $82 \pm 4$ & $92 \pm 1$ & $91 \pm 3 $ &$\mathbf{94 \pm 3^{*}}$\\ \hline 
    2D Overlap Percentage (\%) & $51 \pm 5$ & $46 \pm 6$ & $\mathbf{24 \pm 2^{*}}$ & $27 \pm 2$ & $ 57 \pm 8$ & $51 \pm 9$ & $37 \pm 5$ &$\mathbf{35 \pm 1^{*}}$ & $47 \pm 7$ & $39 \pm 8$  &$\mathbf{24 \pm 8}^{*}$ & ${26 \pm 2}$\\ \hline
    3D Overlap Percentage (\%) & $68 \pm 1$ & $67 \pm 6$ & $65 \pm 5$ & $\mathbf{51 \pm 4^{*}}$ & $ 42 \pm 7$ & $49 \pm 1$ & $38 \pm 2$ &$\mathbf{28 \pm 6^{*}}$ & $40 \pm 13$ & $37 \pm 12$ & $41 \pm 2$ &$\mathbf{19 \pm 3^{*}}$\\ \hline
    3D-Map RMSE & $3.71$ & $3.78$ & $3.68$ & $\mathbf{0.08^{*}}$ & $2.55$ & $2.88$ & $2.79$& $\mathbf{0.12^{*}}$& $4.10$& $3.95$& $3.15$ & $\mathbf{0.07^{*}}$ \\ \hline
    Bounding Volume (\%) & $10.1$ & $9.95$ &  $9.78$&$\mathbf{0.02}^{*}$ & $8.1$ & $7.76$ & $9.64$ &$\mathbf{0.01}^{*}$ & $20.5$& $21.7$& $19.5$ &$\mathbf{0.52}^{*}$\\ 
    \hline
  \end{tabular}}
\end{table*}

\begin{figure}[t]
    \centering
    \captionsetup{font=small}
    \includegraphics[width = \linewidth]{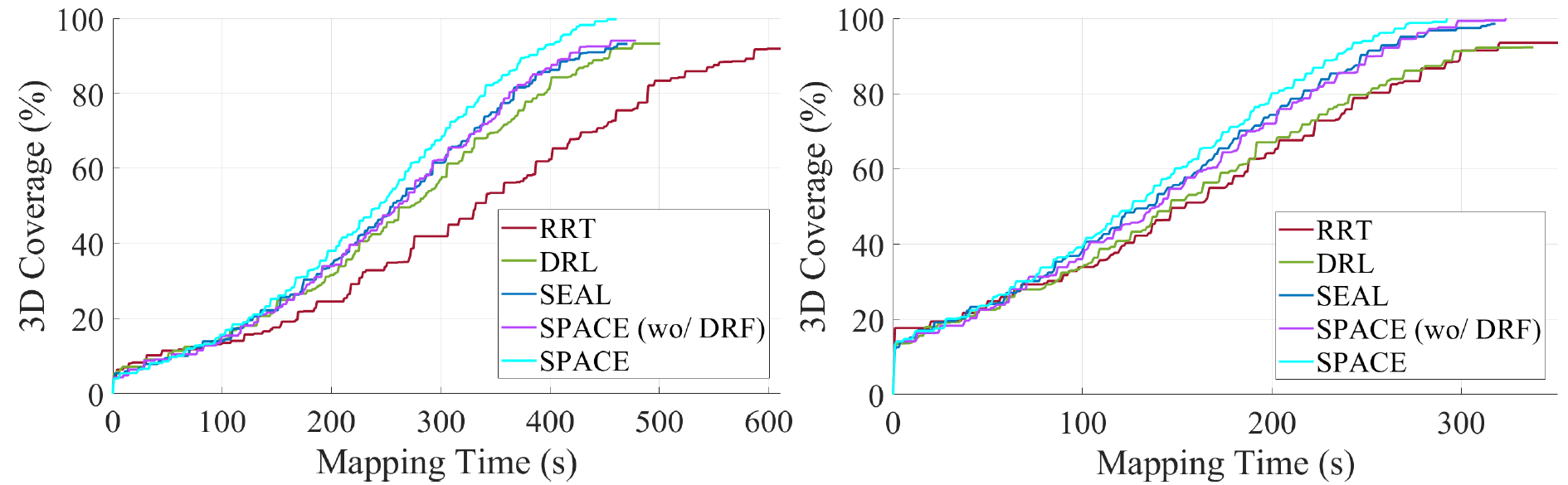}
    \caption{\footnotesize{Performance Analysis of Exploration Strategies with respect to Total Spatial Coverage in tested environments \textbf{(Left)} AWS House World (3 robots) \textbf{(Right)} AWS Bookstore World (3 robots)}.}
    \label{fig:exploration-performance}
    \vspace{-4mm}
\end{figure}

\begin{figure*}[ht]
    \centering
    \captionsetup{font=small}
    \includegraphics[width = 0.99\linewidth]{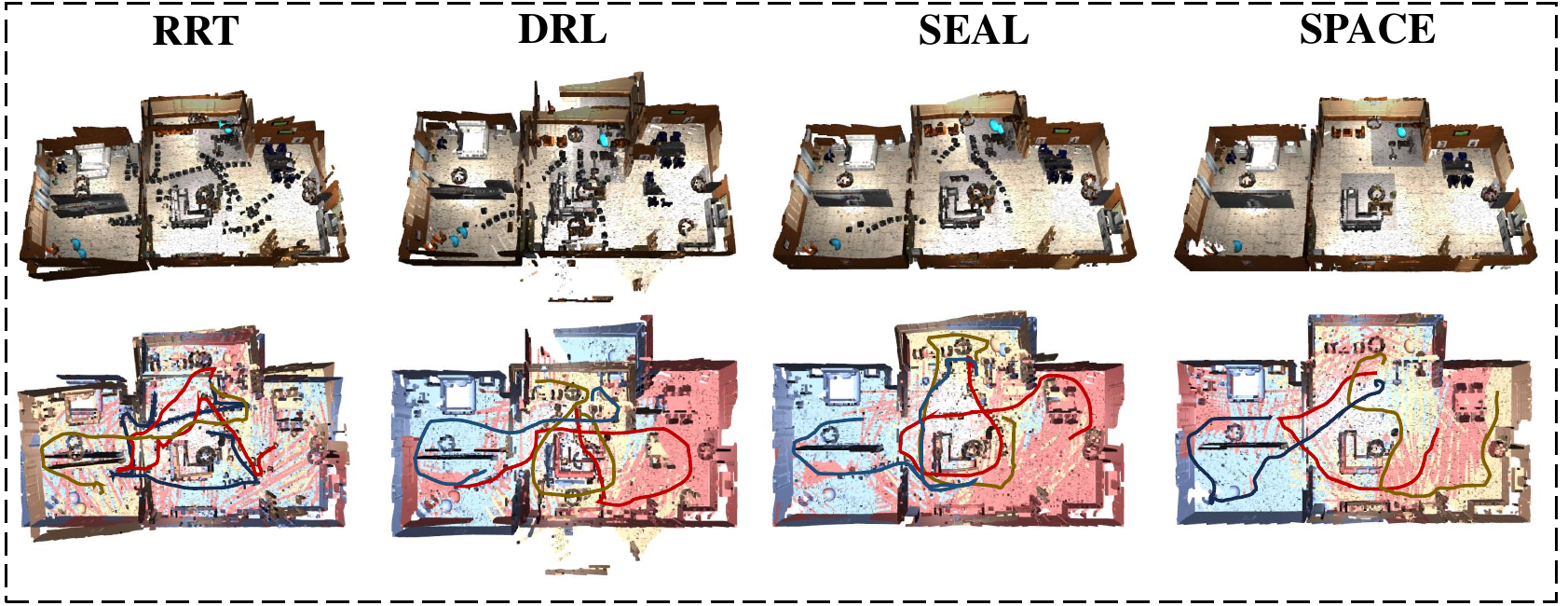}
    \caption{\footnotesize{The 3D Merged and Local Reconstruction Maps with trajectories of each robot from experiments in the AWS House and with 3 robots.}}
    \label{fig:3dmap1}
\end{figure*} 

\begin{figure*}[ht]
    \centering
    \captionsetup{font=small}
    \includegraphics[width = 0.99\linewidth]{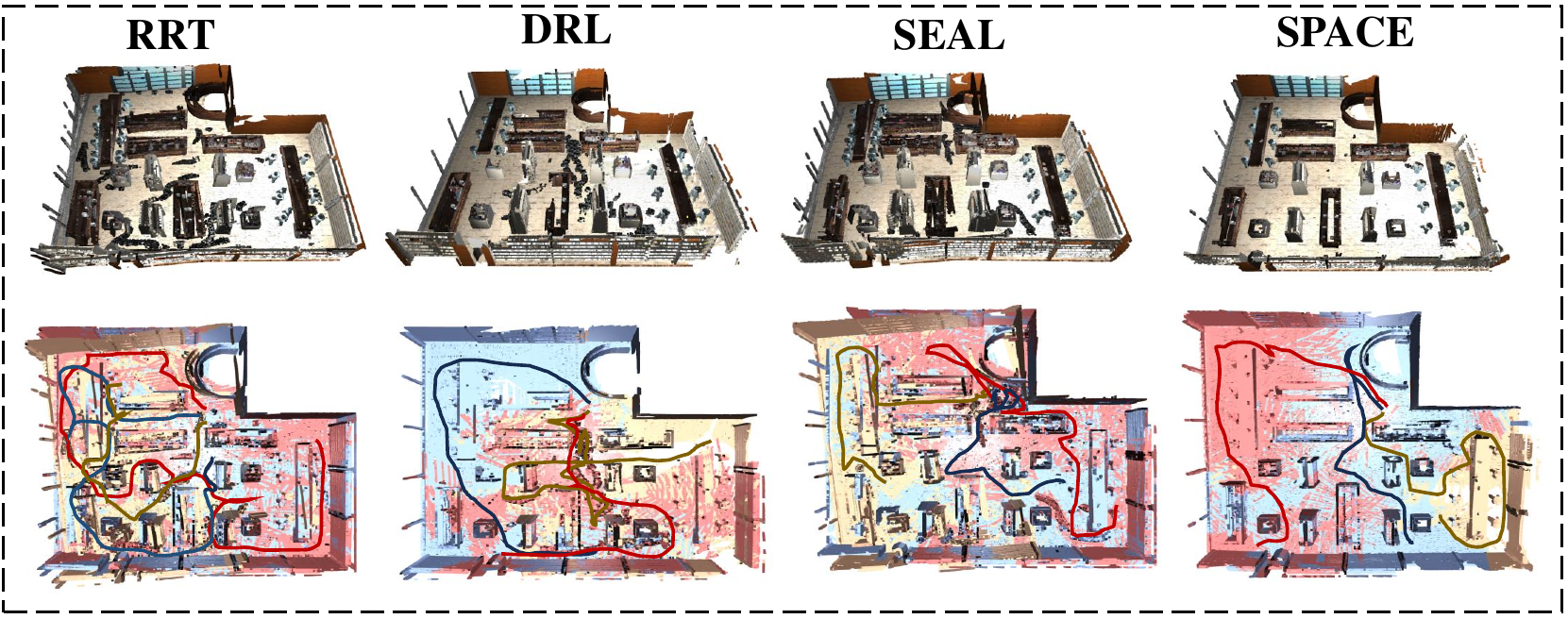}
    \caption{\footnotesize{The 3D Merged and Local Reconstruction Maps with trajectories of each robot from experiments in the AWS Bookstore and with 3 robots.}}
    \label{fig:3dmap2}
\end{figure*} 

\subsection{Exploration Performance}
The SPACE multi-robot exploration approach is compared with the recent architectures RRT \cite{zhang2020rapidly}, DRL \cite{hu2020voronoi}, and SEAL \cite{latif2023seal} in the AWS House and AWS Bookstore worlds. 
The results are presented in Table \ref{tab:explo-com} and an analysis of the efficiency of the exploration approaches with respect to the spatial coverage is depicted in Fig. \ref{fig:exploration-performance}. The SPACE surpasses the other methods, achieving maximum coverage in less time. Moreover, SPACE without DRF performed comparably to SEAL, and both configurations significantly outperformed the RRT and DRL methods.

The spatial explored maps across the AWS House and AWS Bookstore worlds with benchmark exploration algorithms are shown in Fig. \ref{fig:3dmap1}, \ref{fig:3dmap2}. The SPACE consistently outperformed the RRT and DRL in key performance metrics, achieving broader coverage in less time and distance, with an average 5.1\% and 14.3\% increase in 2D and 3D mapping coverage, respectively. Although the mapping time is 1\% higher than the DRL in three robot house worlds, the total path is reduced by almost 8\%. The overall mapping time of SPACE-Map is improved by 5\% compared to the existing architectures. Moreover, we observed a reduction of an average of 26 meters in travel distance compared to the best in RRT, DRL, and SEAL in the conducted experiments. The grid map overlap percentage is reduced by 18\% in average across all the scenarios. Moreover, the 3D overlap percentage outperformed RRT, DRL and SEAL by 20\% in all the scenarios. On average, the RMSE of the SPACE 3D map is approximately 97.85\% lower when compared to the 3D maps explored using RRT, DRL, and SEAL methods. Moreover, the effect of the bounding volume of the ghosting trail is negligible in SPACE.

\section{Conclusion}
We introduced SPACE, a multi-robot spatial exploration pipeline optimized for indoor environments, where robots are colocated during exploration. 
Our semi-distributed approach maximizes the efficiency of dense metric 3D mesh models and their utility for exploration, accounting for the complex spatial constraints within indoor environments. Moreover, SPACE surpassed in exploration and mapping performances compared to other benchmark exploration strategies in extensive simulation experiments. SPACE offers fast spatial exploration, making it beneficial for indoor robot applications.

\bibliographystyle{IEEEtran}
\bibliography{Main}

\end{document}